\pdfoutput=1
\documentclass{article}

\usepackage{arxiv}          %
\usepackage[utf8]{inputenc}
\usepackage[T1]{fontenc}
\usepackage{natbib}
\setcitestyle{authoryear,round,citesep={;},aysep={,},yysep={;}}

\usepackage{amsmath,amsfonts,bm}

\def\eqref#1{equation~\ref{#1}}

\def\1{\bm{1}}

\DeclareMathAlphabet{\mathsfit}{\encodingdefault}{\sfdefault}{m}{sl}
\SetMathAlphabet{\mathsfit}{bold}{\encodingdefault}{\sfdefault}{bx}{n}

\usepackage{url}
\usepackage{amsmath,amssymb}
\usepackage{graphicx}
\usepackage{wrapfig}
\usepackage{booktabs}
\usepackage{multirow}
\usepackage{xcolor}
\usepackage{subcaption}
\usepackage{enumitem}
\usepackage{microtype}
\usepackage{placeins}
\usepackage[colorlinks=true,citecolor=blue,linkcolor=blue,urlcolor=blue]{hyperref}

\newcommand{\method}{StairDrop}
\newcommand{\sprint}{SPRINT}
\newcommand{\tread}{TREAD}

\title{StairDrop: Hierarchical Token Dropping for\\Diffusion Transformers}

\author{%
  Bingshuo Qian \qquad Zeyu Michael Li \qquad Jiayin Liu \qquad Xiang Cheng \\[4pt]
  {\normalfont Duke University} \\
  {\normalfont\texttt{\{bingshuo.qian, zeyu.li030, jiayin.liu2, xiang.cheng\}@duke.edu}}
}
\renewcommand{\shorttitle}{StairDrop: Hierarchical Token Dropping for Diffusion Transformers}
\renewcommand{\undertitle}{}
\date{}

\begin{document}
\maketitle

\begin{abstract}
Diffusion transformers generate high-quality images, but training them is expensive, since every block processes every image token. Token dropping reduces this cost by letting only part of the tokens through the middle of the network during training, and existing methods do so in a single step: tokens are dropped at one depth and restored at another. Inspired by how a U-Net contracts and then expands its resolution, we propose \method{}, which reduces the number of participating tokens step by step as the network goes deeper, keeps a small set of them through a sparse core, and then restores them step by step on the way out, so that participation along depth forms a staircase. Controlled studies on SiT-B/2 show that the intermediate level improves generation, and from them we propose guidelines for dividing depth among the stages, which we follow on SiT-XL/2. On ImageNet $256{\times}256$, \method{} matches the FID of 1.96 reported for the one-level method \sprint{} with classifier-free guidance, using \textbf{half of its training compute}, and with interval guidance it reaches \textbf{FID 1.47}, lower than every token-dropping method we compare with.
\end{abstract}

\section{Introduction}
\label{sec:intro}

\begin{wrapfigure}{r}{6.6cm}
  \vspace{-10pt}
  \centering
  \includegraphics[width=\linewidth,trim=12bp 12bp 12bp 12bp,clip]{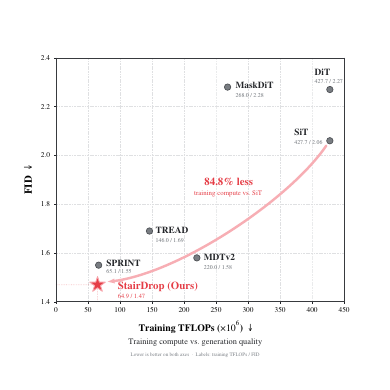}
  \captionsetup{font=small,skip=3pt}
  \caption{\textbf{Training compute and sample quality.} Selected systems from Table~\ref{tab:system}, each at its reported guided sampling setting. Labels give training TFLOPs ($\times 10^6$) and FID-50K on ImageNet $256{\times}256$. The arrow compares with SiT-XL/2.}
  \label{fig:compute-quality}
\end{wrapfigure}

Diffusion transformers have become the standard backbone for image generation, from class-conditional synthesis~\citep{peebles2023dit,ma2024sit} to text-to-image models~\citep{esser2024sd3}. Their training is expensive: every block processes every token of the image at every step, attention grows quadratically with the number of tokens, and SiT-XL/2, for instance, is trained for 7M steps~\citep{ma2024sit}. Token dropping reduces this cost by letting many tokens skip part of the network during training, while the model still uses all tokens at generation time. The idea goes back to masked autoencoders~\citep{he2022mae}, which apply the encoder only to the visible patches, and it has made diffusion transformers several times cheaper to train at comparable image quality~\citep{krause2025tread,park2025sprint}.

Successive methods have changed where tokens are dropped and how they come back. MaskDiT~\citep{zheng2024maskdit} and MDT~\citep{gao2023mdt} withhold tokens from the input onward and recover the missing positions with a decoder or an interpolator. \tread{}~\citep{krause2025tread} and \sprint{}~\citep{park2025sprint} instead keep the first and last blocks dense and drop tokens only in the blocks between them, carrying or fusing the dense features across this sparse span. This lets them drop up to three quarters of the tokens while keeping image quality. Along this line, the choice of which tokens to keep and how to restore the others has become more refined, but participation, the number of tokens a block processes, has kept one form along depth: it falls from all tokens to the reduced set at one block and returns at another, with nothing in between. We refer to this as one-level token dropping.

The U-Nets on which diffusion models were built before transformers~\citep{ho2020ddpm,dhariwal2021adm} move between full and reduced resolution in a different way. A U-Net~\citep{ronneberger2015unet} reaches its coarsest resolution through a series of intermediate ones, processes the features at each of them on the way down, and returns through the same resolutions on the way up, with a skip connection at each step. One-level token dropping can be read as a U-Net with a single step: withholding tokens plays the role of downsampling, the sparse span sits where the bottleneck would be, and restoring the tokens and fusing them with the features saved before the span plays the role of upsampling and the skip connection. What it lacks are the intermediate scales. Their counterpart in token dropping is an intermediate level of participation, at which blocks process more tokens than the sparse span and fewer than the full sequence, once on the way down and once on the way up.

\method{} adds these intermediate levels to token dropping (Figure~\ref{fig:method}). On the way down, a first selection keeps half of the tokens and a second selection keeps half of those, so that participation falls to one half and then to one quarter of the full sequence. On the way up, the levels are restored in reverse order, each with a skip fusion, and participation along depth forms a staircase. The levels also make the depth of the network a design choice: a backbone with a fixed number of blocks has to divide them among the dense ends, the intermediate levels and the sparse core, and we study how this division affects generation. The levels exist only during training. At generation time every block processes every token, and the network is the standard dense transformer with two linear fusion layers.

We first study \method{} in controlled comparisons on SiT-B/2. Added to a one-level model with the same blocks, fusions and parameters, the intermediate level lowers FID-10K from 37.85 to \textbf{29.45}, and varying the division of depth one stage at a time yields three guidelines, which we follow to build the layouts for SiT-XL/2. As a complete training system on ImageNet $256{\times}256$ (Figure~\ref{fig:compute-quality}), \method{} matches the FID of 1.96 that \sprint{} reports with classifier-free guidance using \textbf{half of its training compute}, and with interval guidance it reaches \textbf{FID 1.47}, lower than every token-dropping method we compare with. It also combines with representation alignment~\citep{yu2025repa}, and the gains of the two methods add. Its gains hold on two further backbones and on text-to-image generation as well. Finally, Section~\ref{sec:understanding} analyzes the trained models: as in a U-Net, the range over which tokens interact widens level by level on the way down and narrows level by level on the way up, while one-level dropping changes it in a single step.

\section{Related Work}
\label{sec:related}

\paragraph{Efficient training through token participation.}
Masked autoencoders~\citep{he2022mae} reduce the cost of visual representation learning by applying the encoder only to the visible patches. MaskDiT~\citep{zheng2024maskdit} brings masked training to diffusion models with an asymmetric encoder--decoder and an auxiliary reconstruction objective. MDT~\citep{gao2023mdt} uses masked latent modeling and a side interpolator that recovers the missing positions, and MicroDiT~\citep{sehwag2024microdit} mixes patch features before masking. Routing methods instead preserve the information of the tokens that skip computation: \tread{}~\citep{krause2025tread} carries their features across the routed blocks, and \sprint{}~\citep{park2025sprint} fuses the sparse output with the dense features that entered the span. These methods show that the number of participating tokens is an effective way to control training cost. All of them use one level of participation. \method{} adds intermediate levels and studies how to divide the blocks of the backbone among them. Within each level it uses \sprint{}'s placeholder restoration and linear fusion, and Appendix~\ref{app:restore} compares alternative restorations.

\paragraph{Hierarchical and U-shaped architectures.}
U-Net~\citep{ronneberger2015unet} combines progressive spatial contraction and expansion with skip connections between corresponding scales. Diffusion transformers adopt different parts of this design. U-ViT~\citep{bao2023uvit} connects shallow and deep blocks through long skips without spatial downsampling. HDiT~\citep{crowson2024hdit} builds a spatial hierarchy by merging neighboring tokens and splitting them again during expansion, and U-DiT~\citep{tian2024udit} combines a U-shaped backbone with downsampled self-attention. UDT~\citep{yun2026udt} merges and unmerges tokens progressively according to feature similarity. Outside diffusion, fractional max-pooling~\citep{graham2014fractional} reduces spatial resolution gradually, and DynamicViT~\citep{rao2021dynamicvit} prunes tokens progressively for efficient classification. These architectures change the resolution of the tokens in the network that generates images. \method{} instead changes only how many tokens take part, and only during training: tokens keep their coordinates and feature width while nested subsets are selected and restored, and at generation time every token passes through every block.

\paragraph{Complementary training and sampling methods.}
Representation alignment is a separate route to more efficient diffusion training. REPA~\citep{yu2025repa} adds an auxiliary learning signal by aligning projected hidden states with clean-image features from pretrained visual encoders. \method{} changes which tokens each stage processes and leaves the objective unchanged, so the two can be combined (Section~\ref{sec:exp-general}). The sampler also affects the quality of a trained model. Classifier-free guidance~\citep{ho2022cfg} and guidance intervals~\citep{kynkaanniemi2024interval} modify the denoising trajectory, and Sparse Guidance~\citep{sparseguidance2026} builds a guidance signal from token sparsity.

\section{\method{}}
\label{sec:method}

\begin{figure}[t]
  \centering
  \includegraphics[width=\linewidth]{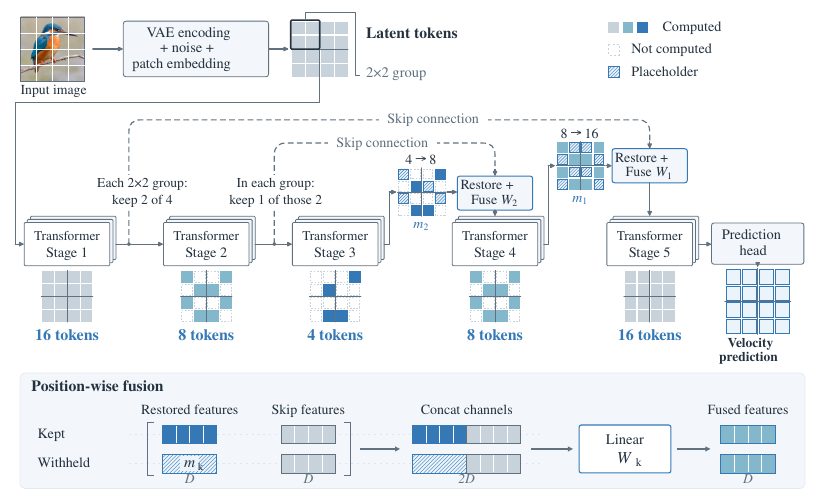}
  \caption{\textbf{\method{} during training.} The 256-token model drawn with 16 positions. The first selection keeps two positions of every $2{\times}2$ group and the second keeps one of those two, so the five stages see 16, 8, 4, 8 and 16 tokens. Restoration runs from the inside out: withheld positions receive the placeholder $\mathbf{m}_k$ of their level, and the dashed skips bring the features from before each selection to the fusion $W_k$. \textbf{Bottom:} the fusion concatenates each position's computed feature or placeholder with its skip feature along channels and applies a linear layer, shared across the positions of a level.}
  \label{fig:method}
\end{figure}

\subsection{Preliminaries}
\label{sec:prelim}

\paragraph{Flow matching.}
We work with SiT~\citep{ma2024sit}, which is trained as a flow-matching model on the linear interpolant between a data latent $x_0$ and Gaussian noise $\varepsilon$,
\begin{equation}
  x_t = (1-t)\,x_0 + t\,\varepsilon, \qquad t \in [0, 1],
  \label{eq:interpolant}
\end{equation}
so that $t = 1$ is the noise end. The network $v_\theta(x_t, t, c)$ is a stack of $L$ identical Transformer blocks conditioned on the time $t$ and the class label $c$. It is trained to predict the velocity of the interpolant by minimizing
\begin{equation}
  \mathcal{L}(\theta) = \mathbb{E}_{x_0,\,\varepsilon,\,t}\,\big\|\,v_\theta(x_t, t, c) - (\varepsilon - x_0)\,\big\|^2 ,
  \label{eq:loss}
\end{equation}
and samples are drawn by integrating the learned velocity from $t = 1$ to $t = 0$ with an ODE or an SDE sampler. During sampling, classifier-free guidance (CFG)~\citep{ho2022cfg} replaces the conditional velocity with an extrapolation away from the unconditional velocity $v_\theta(x_t, t, \varnothing)$, with guidance scale $w$:
\begin{equation}
  \tilde v_\theta(x_t, t, c) = v_\theta(x_t, t, \varnothing) + w\,\big(v_\theta(x_t, t, c) - v_\theta(x_t, t, \varnothing)\big), \qquad w > 1 .
  \label{eq:cfg}
\end{equation}

\paragraph{One-level token dropping.}
The latent is split into $N$ patches, the token positions, each represented by a feature of width $D$. With patch size 2 on a $32{\times}32$ latent, this gives $N = 256$ positions on a $16{\times}16$ grid. One-level dropping runs a span of consecutive blocks on a subset $S$ of these positions, with every position present before and after the span. The subset is drawn anew for each sample at each training step. Let $h_{\mathrm{in}}\in\mathbb{R}^{N\times D}$ be the state that enters the span and $h_S$ the output of the span on the kept positions. Restoration brings the sequence back to $N$ positions and combines it with the saved entry state,
\begin{equation}
  h \;\leftarrow\; F\big(\,\mathrm{restore}(h_S;\,p)\,,\; h_{\mathrm{in}}\big),
  \label{eq:fusion}
\end{equation}
where $\mathrm{restore}(h_S;p)$ places $h_S$ at its original positions and fills the withheld positions with $p$, and the fusion $F$ merges this sequence with $h_{\mathrm{in}}$ position by position. \tread{}~\citep{krause2025tread} keeps a uniformly sampled half of the positions, fills the withheld ones with their own entry features from $h_{\mathrm{in}}$, and uses no fusion. \sprint{}~\citep{park2025sprint} keeps one position of every $2{\times}2$ cell of the grid and fills the others with a fixed placeholder vector $\mathbf{m}\in\mathbb{R}^{D}$. Its fusion concatenates the two sequences along channels and applies a linear layer shared across positions, $F(u,h_{\mathrm{in}})=W[u;h_{\mathrm{in}}]+b$ with $W\in\mathbb{R}^{D\times2D}$ and $b\in\mathbb{R}^{D}$. \method{} uses cell-based selection and this placeholder-and-fusion restoration at each of its levels.

\subsection{\method{}}
\label{sec:staircase}

\paragraph{The staircase.}
Let $S_0$ be the full set of $N$ token positions. \method{} runs the blocks of the backbone on a nested sequence of position sets $S_0 \supset S_1 \supset \dots \supset S_K$, with $N_k = |S_k|$ tokens at level $k$ (Figure~\ref{fig:method}). On the way down, a group of blocks runs at each level before the next selection, the core runs at the innermost level $K$, and on the way up the levels are restored in reverse order, again with a group of blocks at each level. The number of tokens a block processes thus falls level by level to $N_K$ and returns the same way, forming a staircase along depth. One-level dropping is the case $K = 1$. \method{} uses $K = 2$ with $N_1 = N/2$ and $N_2 = N/4$, which gives five stages: $d_0$ dense blocks, $d_1$ blocks on $N/2$ tokens, a core of $d_c$ blocks on $N/4$ tokens, $d_2$ blocks on $N/2$ tokens and $d_3$ dense blocks, written as the layout $d_0/d_1/d_c/d_2/d_3$.

\paragraph{Selection.}
Each step down keeps a random half of the tokens present, chosen within the $2{\times}2$ cells of the token grid: the first step keeps two of the four positions of every cell, and the second keeps one of those two. Halving within cells mirrors the $2{\times}2$ downsampling of a U-Net: every cell stays represented at every level, so each level covers the image evenly. Each set is a subset of the one before it, so a position withheld at one level stays withheld until that level is restored. Kept tokens retain their position embeddings, and attention at a level is ordinary attention among its tokens. The core keeps one position per cell, the same density as \sprint{}'s sparse span, so \method{} keeps the core as sparse as one-level dropping and adds the intermediate level around it. Table~\ref{tab:split} compares other ways of splitting the same reduction, including unequal ratios and a third level.

\paragraph{Restoration.}
Each step up undoes the matching step down using the operations of Section~\ref{sec:prelim}. At the end of level $k$, the positions of $S_{k-1}$ that are not in $S_k$ receive the placeholder $\mathbf{m}_k$ of that level, and the sequence is fused with the state $h^{(k)}_{\mathrm{in}}$ that entered the level,
\begin{equation}
h \;\leftarrow\; W_k\big[\,\mathrm{restore}(h;\mathbf{m}_k)\,;\,h^{(k)}_{\mathrm{in}}\,\big] + b_k ,
\label{eq:level-fusion}
\end{equation}
the fusion of Eq.~(\ref{eq:fusion}) with a separate linear layer for each level. The levels are restored from the inside out, so each skip connects the two sides of the same level, as the skip connections of a U-Net connect the two sides of the same resolution. Each position therefore leaves a level carrying either its own output at that level or the placeholder, together with its state from before the level. As in one-level dropping, the loss is computed on every position.

\paragraph{Generation.}
No token is withheld at generation time: every block processes every token and each fusion reads the full sequence, so the model that generates images is the dense backbone with one linear fusion layer per level, two for \method{}.

\subsection{Dividing the depth}
\label{sec:allocation}

The levels make the depth of the network a design variable. A one-level layout only chooses the depths of its two dense ends, and the sparse span takes the remaining blocks. A layout with $K$ levels chooses the depths of $2K+1$ stages that sum to $L$, with at least one block each. For \method{} these are $d_0$, $d_1$, $d_c$, $d_2$ and $d_3$, and four of them are free. The layout also sets the cost of a training step. Counting each block in proportion to the number of tokens it processes, a step costs, relative to a dense step,
\begin{equation}
\frac{1}{L}\sum_{s} d_s\,\frac{N_{k(s)}}{N} \;=\; \frac{(d_0 + d_3) + \tfrac{1}{2}(d_1 + d_2) + \tfrac{1}{4}\,d_c}{L} ,
\label{eq:cost}
\end{equation}
where the sum runs over the stages and $k(s)$ is the level of stage $s$.

\section{Experiments}
\label{sec:experiments}

\paragraph{Setup.}
All experiments except the text-to-image study of Section~\ref{sec:exp-backbones} are class-conditional ImageNet $256{\times}256$~\citep{deng2009imagenet} in the latent space of the Stable Diffusion VAE~\citep{rombach2022ldm}. Samples are scored with the ADM evaluator~\citep{dhariwal2021adm} against its ImageNet reference batch, which reports FID~\citep{heusel2017fid}, Inception score~\citep{salimans2016is}, and precision and recall~\citep{kynkaanniemi2019pr}. We write FID-10K and FID-50K for FID on 10K and 50K samples.

The experiments use two protocols. The design study on SiT-B/2 and the scaling study on SiT-XL/2 (Sections~\ref{sec:exp-b2} and~\ref{sec:exp-xl}) use the original SiT backbone and recipe~\citep{ma2024sit}. Every arm is trained from scratch on the same latents in the same order. It is evaluated on its sparse weights, before any dense stage, with the 50-step Euler ODE sampler of \sprint{}'s protocol (ODE-50)~\citep{park2025sprint} and without guidance, except in the last column of Table~\ref{tab:xl400}. The system-level comparison (Section~\ref{sec:exp-system}) adopts \sprint{}'s published setting in full, including its improved backbone, its dense stage and its guided sampling, as described there. We re-implement \tread{} and \sprint{} from their released code in a single code base, so that all methods share the backbone, the recipe, the data order and the evaluator. Recipes, the dense stage, sampling and compute accounting are detailed in Appendices~\ref{app:protocol} and~\ref{app:compute}.

\subsection{Design study on SiT-B/2}
\label{sec:exp-b2}

\begin{figure}[t]
  \centering
  \includegraphics[width=\linewidth]{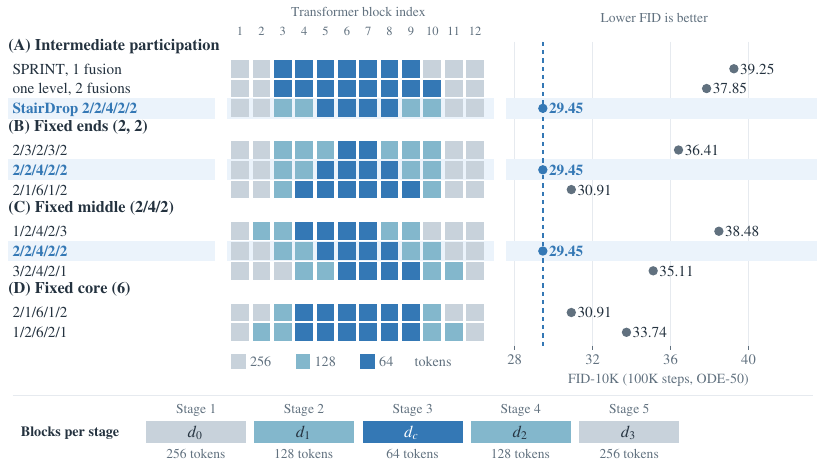}
  \caption{\textbf{Design study on SiT-B/2.} FID-10K at 100K steps (ODE-50, no guidance). Each square is a block, colored by the number of tokens it processes during training, and layouts are written $d_0/d_1/d_c/d_2/d_3$. (A) Intermediate participation: one-level dropping with \sprint{}'s mechanism over blocks 3--9 and one fusion, a one-level control with the span, fusion positions and parameters of \method{}, and \method{} $2/2/4/2/2$. (B)--(D) Depth allocation with one quantity varied at a time: dense ends fixed (B), the $d_1/d_c/d_2$ middle fixed (C), core fixed (D). The default $2/2/4/2/2$ is highlighted wherever it appears.}
  \label{fig:layouts}
\end{figure}

The design study uses SiT-B/2, with twelve blocks, trained for 100K steps and evaluated under the ODE-50 protocol (Figure~\ref{fig:layouts}; Appendix~\ref{app:full} lists every arm with its other metrics). It first tests the intermediate level with the parameters and the depth held fixed, and then studies how a fixed budget of blocks should be divided among the stages.

\paragraph{The intermediate level.}
Our reference is one-level dropping with \sprint{}'s mechanism over blocks 3--9 and one fusion (Figure~\ref{fig:layouts}A), which reaches FID-10K 39.25. Compared with this reference, \method{} adds a second placeholder and fusion, and its sparse region extends to block 10. To separate these additions from the intermediate level itself, we build a one-level control with the span, the fusion positions and the parameters of \method{}. It keeps 64 tokens through blocks 3--10, adds a second placeholder and fusion at the depths where \method{} restores, and reaches 37.85. \method{} $2/2/4/2/2$ differs from this control only in blocks 3--4 and 9--10, which process 128 tokens instead of 64, and it reaches 29.45. Since the control has the same added parameters, the gain from 37.85 to 29.45 comes from the intermediate level of participation.

\paragraph{Allocating depth across stages.}
\label{sec:exp-alloc}
With the intermediate level in place, the twelve blocks can be divided among the five stages in many ways. We vary one quantity at a time around $2/2/4/2/2$ (Figure~\ref{fig:layouts}B--D). With the dense ends fixed at two blocks, shrinking each intermediate stage from three blocks to two (the core growing from two blocks to four) lowers FID-10K from 36.41 to 29.45, and shrinking it to one (core of six blocks) raises it to 30.91 (B). With the middle fixed, an even split of the dense blocks is ahead of both uneven splits, and the input end is the more sensitive of the two (C). With the core fixed, the remaining depth is better spent at the dense ends than at the intermediate stages (D).

Among the tested layouts, $2/2/4/2/2$ gives the lowest FID, and we use it as the SiT-B/2 default from here on. What we carry to the deeper backbone is not this layout but three guidelines: the core should be at least as deep as either intermediate stage, the intermediate stages and the dense ends should have at least two blocks each, and the two dense ends should have equal depth. Section~\ref{sec:exp-xl} tests them on SiT-XL/2.

\paragraph{Splitting the reduction.}
\method{} splits the reduction of one-level dropping, from 256 to 64 tokens, into two equal halvings. Table~\ref{tab:split} compares other ways of splitting the same reduction, all with the dropped span in blocks 3--9: one level, a deeper level of 32 or 16 tokens inside the core, and unequal ratios with a first level of 192 tokens. Two equal halvings are ahead of every other split, at a lower step cost than the 192-token first level, and a deeper level inside the core improves on one level by at most 1.1.

\begin{table}[t]
  \centering\small
  \caption{\textbf{Splitting the reduction from 256 to 64 tokens.} SiT-B/2 with the dropped span in blocks 3--9, 100K steps, FID-10K with SDE-250, no guidance.}
  \label{tab:split}
  \begin{tabular}{@{}lcr@{}}
    \toprule
    Split & Tokens in the span & FID-10K \\
    \midrule
    One level & 64 & 36.03 \\
    Deeper level in the core, blocks 5--6 & 64 / 32 / 64 & 34.94 \\
    Deeper level in the core, blocks 5--7 & 64 / 32 / 64 & 38.37 \\
    Deeper level in the core, blocks 5--7 & 64 / 16 / 64 & 38.89 \\
    Unequal ratios, $4/3{\times}$ then $3{\times}$ & 192 / 64 / 192 & 35.14 \\
    Two equal halvings (\method{}) & 128 / 64 / 128 & \textbf{31.95} \\
    \bottomrule
  \end{tabular}
\end{table}

\paragraph{Longer training.}
Table~\ref{tab:b2long} follows the dense, \tread{}, one-level and \method{} models to 400K steps, scored on 50K samples. Their ranking at 100K steps still holds at 400K, and \method{} ends 3.8 FID below one-level dropping. Appendix~\ref{app:equal-compute} compares these models at equal training compute.

\begin{table}[t]
  \small
  \begin{minipage}[t]{0.38\linewidth}
    \centering
    \caption{\textbf{SiT-B/2.} ODE-50, no guidance.}
    \label{tab:b2long}
    \setlength{\tabcolsep}{4pt}
    \begin{tabular}{@{}lrr@{}}
      \toprule
      & FID-10K & FID-50K \\
      Model & 100K & 400K \\
      \midrule
      Dense & 70.72 & 38.58 \\
      \tread{} & 54.71 & 20.58 \\
      One-level & 39.25 & 11.48 \\
      \method{} & \textbf{29.45} & \textbf{7.66} \\
      \bottomrule
    \end{tabular}
  \end{minipage}\hfill
  \begin{minipage}[t]{0.59\linewidth}
    \centering
    \caption{\textbf{SiT-XL/2.} Sparse weights, ODE-50.}
    \label{tab:xl400}
    \setlength{\tabcolsep}{4pt}
    \begin{tabular}{@{}llrrrr@{}}
      \toprule
      & & \multicolumn{3}{c}{FID-10K, no guidance} & Guided \\
      \cmidrule(lr){3-5}\cmidrule(l){6-6}
      Model & Layout & 200K & 300K & 400K & 400K \\
      \midrule
      One-level & $2/24/2$ & 11.52 & 9.65 & 8.75 & 7.90 \\
      \method{} & $2/2/20/2/2$ & 9.74 & 8.47 & 8.08 & 8.43 \\
      \method{} & $3/3/16/3/3$ & 11.70 & 8.46 & \textbf{7.52} & 7.48 \\
      \method{} & $4/6/8/6/4$ & 11.20 & 8.76 & 7.68 & \textbf{7.12} \\
      \bottomrule
    \end{tabular}
  \end{minipage}
\end{table}

\subsection{Scaling to SiT-XL/2}
\label{sec:exp-xl}

Table~\ref{tab:xl400} repeats the controlled comparison on SiT-XL/2, which has 28 blocks instead of 12, with three \method{} layouts built from these guidelines. The one-level baseline is \sprint{}'s XL layout, trained with the same recipe and evaluated under the same protocol. The layout $2/2/20/2/2$ is this baseline with the two core blocks nearest each dense end run on 128 tokens instead of 64, and it is ahead of the baseline at every measured step. At 400K steps $3/3/16/3/3$, the lowest of the three, is 1.2 FID-10K below one-level dropping at the same recall, and 4.76 against 6.04 on 50K samples. Appendix~\ref{app:equal-compute} lists these runs by training compute.

We choose the XL layout for the system-level comparison among these three. The thin-ended $2/2/20/2/2$ leads at 200K steps, and the two layouts with thicker dense ends overtake it by 300K to 400K steps. Since the system-level comparison is made with guidance, we also evaluate the layouts with guidance (last column of Table~\ref{tab:xl400}; CFG on $t \le 0.7$ at the scale tuned for each layout). This evaluation uses the sparse weights, before the dense stage. Sparse training changes how a model responds to guidance: MaskDiT, \tread{} and \sprint{} all end with a dense stage before guided sampling, and \tread{} notes that restricting information during training can degrade performance under guidance~\citep{zheng2024maskdit,krause2025tread,park2025sprint}. We therefore use this column only to choose the layout. With guidance, $4/6/8/6/4$ reverses the unguided order and reaches the lowest FID, 7.12 against 7.48 for $3/3/16/3/3$, 8.43 for $2/2/20/2/2$ and 7.90 for one-level dropping. We take $4/6/8/6/4$ as the XL layout and continue it on the sparse schedule to 800K steps (Appendix~\ref{app:full}). Every system-level \method{} result below uses this layout.

\subsection{System-level comparison}
\label{sec:exp-system}

\begin{figure}[t]
  \centering
  \includegraphics[width=\linewidth]{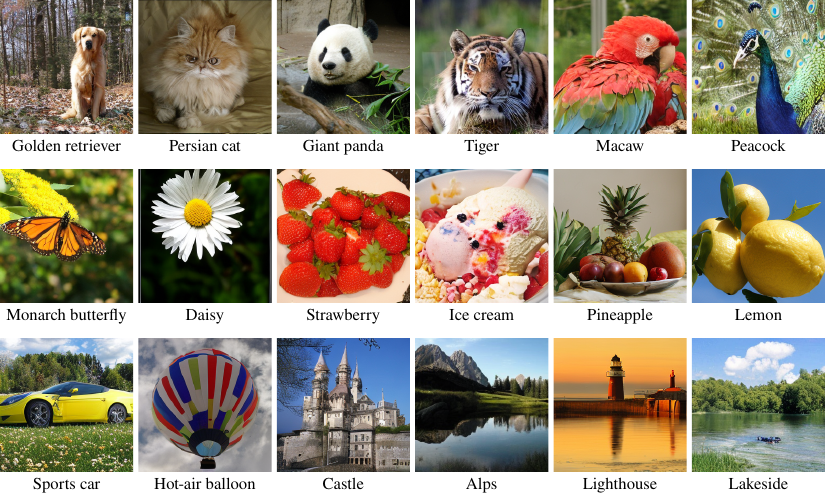}
  \caption{\textbf{Selected ImageNet samples from \method{}.} One $256{\times}256$ image from each of 18 categories, generated by the SiT-XL/2 model with FID-50K 1.47 in Table~\ref{tab:system} (SDE-250, interval CFG with $w=1.6$).}
  \label{fig:qualitative-main}
\end{figure}

\begin{table}[t]
  \centering\small
  \caption{\textbf{System-level comparison on ImageNet $256{\times}256$ with guidance, 50K samples.} Published rows carry the numbers of their papers under their own sampling protocols, and \sprint{}'s rows are quoted from its paper. Our rows use the SDE sampler with 250 steps. The \method{} rows are $4/6/8/6/4$ after 800K, 1.72M and (with REPA) 400K sparse steps, each followed by a 20K-step dense stage at batch 1024 (Appendix~\ref{app:protocol}). Epochs are images seen divided by 1.28M, with every training stage counted at its batch size; \sprint{}'s rows therefore include the 100K-step dense stage at batch 1024 that its own table omits, and MaskDiT's count only the masked stage. Training compute is \sprint{}'s measurement, the forward FLOPs of the training graph at batch 256 times the number of iterations, scaled to each paper's training length (Appendix~\ref{app:compute}). $\dagger$: improved SiT backbone from \sprint{}. $\ddagger$: interval or scheduled guidance. PDG: path-drop guidance. --: not available.}
  \label{tab:system}
  \setlength{\tabcolsep}{3.5pt}\footnotesize
  \renewcommand{\arraystretch}{1.15}
  \begin{tabular*}{\linewidth}{@{\extracolsep{\fill}}lccccccc@{}}
    \toprule
    \multicolumn{1}{c}{Method} & Epochs & Params & \begin{tabular}[c]{@{}c@{}}Training TFLOPs\\($\times 10^6$)\end{tabular} & FID-50K & IS & Prec. & Rec. \\
    \midrule
    ADM-G,U~\citep{dhariwal2021adm} & -- & 673M & -- & 3.94 & 215.8 & 0.83 & 0.53 \\
    LDM-4~\citep{rombach2022ldm} & 167 & 400M & -- & 3.60 & 247.7 & 0.87 & 0.48 \\
    U-ViT-H/2~\citep{bao2023uvit} & 400 & 501M & -- & 2.29 & 263.9 & 0.82 & 0.57 \\
    DiT-XL/2~\citep{peebles2023dit} & 1400 & 675M & 427.7 & 2.27 & 278.2 & 0.83 & 0.57 \\
    FiTv2-XL/2~\citep{wang2024fitv2} & 400 & 671M & -- & 2.26 & 261.0 & 0.81 & 0.59 \\
    SiT-XL/2~\citep{ma2024sit} & 1400 & 675M & 427.7 & 2.06 & 277.5 & 0.83 & 0.59 \\
    \midrule
    MDTv2-XL/2$^\ddagger$~\citep{gao2024mdtv2} & 920 & 742M & 220.0 & 1.58 & 314.7 & 0.79 & 0.65 \\
    MaskDiT~\citep{zheng2024maskdit} & 1600 & 730M & 268.0 & 2.28 & 276.6 & 0.80 & 0.61 \\
    \tread{}$^\ddagger$~\citep{krause2025tread} & 740 & 675M & 146.0 & 1.69 & 292.7 & 0.81 & 0.63 \\
    \midrule
    SiT-XL/2$^\dagger$ & 400 & 675M & 122.2 & 2.04 & -- & 0.82 & 0.60 \\
    \;+ \sprint{}, CFG~\citep{park2025sprint} & 460 & 677M & 65.1 & 1.96 & -- & 0.80 & 0.61 \\
    \;+ \sprint{}, PDG$^\ddagger$ & 460 & 677M & 65.1 & 1.55 & -- & 0.80 & 0.64 \\
    \;+ \sprint{}, PDG$^\ddagger$, REPA & 460 & 677M & 66.7 & 1.49 & -- & 0.81 & 0.64 \\
    \midrule
    SiT-XL/2$^\dagger$ + \method{} & 176 & 680M & 32.8 & \textbf{1.96} & 248.9 & 0.81 & 0.61 \\
    SiT-XL/2$^\dagger$ + \method{}$^\ddagger$ & 360 & 680M & 64.9 & \textbf{1.47} & 307.2 & 0.79 & 0.65 \\
    SiT-XL/2$^\dagger$ + \method{} + REPA$^\ddagger$ & 96 & 680M & 19.2 & \textbf{1.58} & 292.1 & 0.80 & 0.64 \\
    \bottomrule
  \end{tabular*}
\end{table}

\paragraph{Protocol.}
To compare with published numbers, we adopt \sprint{}'s setting in full. The backbone is the improved SiT-XL/2 of its paper, which adds query-key normalization~\citep{dehghani2023vit22b,esser2024sd3}, two-dimensional rotary position embeddings~\citep{lu2024fit} and logit-normal timestep sampling~\citep{esser2024sd3}. \method{} also works with these changes: at 400K sparse steps, unguided, $4/6/8/6/4$ reaches FID-10K 6.76 on this backbone against 7.68 on the original one. Training consists of the sparse schedule followed by a dense stage with every position kept, as in MaskDiT, MicroDiT, \tread{} and \sprint{}~\citep{zheng2024maskdit,sehwag2024microdit,krause2025tread,park2025sprint}. Ours lasts 20K steps at batch 1024, where \sprint{}'s lasts 100K. Our guided results use the SDE sampler of SiT~\citep{ma2024sit} with 250 steps (SDE-250) and are scored on 50K samples. The 800K-step row applies CFG~\citep{ho2022cfg} along the whole trajectory. Rows marked $\ddagger$ apply it only for $t \le 0.7$, a one-sided version of the guidance interval of \citet{kynkaanniemi2024interval}. We count compute as \sprint{} does: the forward FLOPs of the training graph at batch 256 times the number of iterations (Appendix~\ref{app:compute}).

\paragraph{Results.}
Table~\ref{tab:system} gives the comparison, and Figure~\ref{fig:compute-quality} plots its main rows. After 800K sparse steps and the dense stage, \method{} reaches the FID-50K of 1.96 that \sprint{} reports with CFG, using $32.8\times10^6$ training TFLOPs where \sprint{} uses $65.1\times10^6$. With sparse training extended to 1.72M steps, the same 20K-step dense stage and interval guidance, \method{} reaches 1.47 with $64.9\times10^6$ TFLOPs. Over three sampling seeds, these two rows give $1.958 \pm 0.003$ and $1.471 \pm 0.001$. \sprint{} with scheduled path-drop guidance (PDG) reaches 1.55 with $65.1\times10^6$ TFLOPs, and 1.49 with $66.7\times10^6$ when REPA is added. The last row adds REPA to \method{} and is discussed in Section~\ref{sec:exp-general}. Figure~\ref{fig:qualitative-main} shows samples of the 1.47 model from its evaluation set, and Appendix~\ref{app:qualitative} shows more.

\subsection{Compatibility with representation alignment}
\label{sec:exp-general}

\method{} changes only which positions a block processes during training, so it can be combined with methods that change the training objective. We test this with REPA~\citep{yu2025repa}, which aligns the state after block 8 with DINOv2 features~\citep{oquab2024dinov2} and reports a large gain on SiT-XL/2. \sprint{} also reports combining REPA with one-level dropping~\citep{park2025sprint}. We attach REPA at the same block. On $4/6/8/6/4$ this is the fourth block of the first intermediate stage, so we align the 128 tokens present there, each at its own position, with coefficient 0.5. At 100K steps, unguided, REPA lowers the FID-10K of \method{} from 13.11 to 8.36 without changing the time per step. At 400K steps the gain is smaller, from 6.76 to 5.98, as REPA also reports for dense SiT~\citep{yu2025repa}. With guidance, after the same dense stage with the alignment kept, the combination reaches FID-50K 1.58 at 96 epochs and $19.2 \times 10^{6}$ TFLOPs (last row of Table~\ref{tab:system}). This is below the 1.96 that plain \method{} reaches with $32.8\times10^6$ TFLOPs, so the gains of the two methods add.

\subsection{Other backbones and text-to-image generation}
\label{sec:exp-backbones}

We train two further backbones at B/2 scale and a text-to-image model, each for 100K steps with dense training, one-level dropping and \method{}, keeping the $2/2/4/2/2$ layout (Table~\ref{tab:backbones}). The first adds to SiT the query-key normalization, rotary position embeddings and logit-normal timestep sampling of the improved backbone of Section~\ref{sec:exp-system}, and the second is LightningDiT~\citep{yao2025lightningdit}. The text-to-image model is SiT-B/2 conditioned on CLIP ViT-L/14 text features~\citep{radford2021clip} through cross-attention, trained on MS-COCO 2014 training images and captions~\citep{lin2014coco} and evaluated against 10K validation images. \method{} improves on one-level dropping in every setting, by 4.9 to 8.2 FID-10K on the three new ones.

\begin{table}[t]
  \centering\small
  \caption{\textbf{Other backbones and text-to-image generation.} B/2 scale, 100K steps, layout $2/2/4/2/2$; FID-10K, ODE-50, no guidance.}
  \label{tab:backbones}
  \begin{tabular}{@{}lrrr@{}}
    \toprule
    Setting & Dense & One-level & \method{} \\
    \midrule
    ImageNet, SiT & 70.72 & 39.25 & \textbf{29.45} \\
    ImageNet, SiT with QK-norm, RoPE and logit-normal sampling & 53.68 & 24.63 & \textbf{19.74} \\
    ImageNet, LightningDiT & 51.12 & 29.87 & \textbf{21.64} \\
    MS-COCO, text-to-image & 31.12 & 24.06 & \textbf{19.12} \\
    \bottomrule
  \end{tabular}
\end{table}

\section{Understanding \method{}}
\label{sec:understanding}

Section~\ref{sec:experiments} shows that an intermediate level of participation improves generation, including against a control with the same blocks, fusions and parameters. This section looks inside the trained models to explain why. We measure how far each stage of the network reaches across the image, and compare how this reach changes along depth with one level of participation and with the staircase.

\paragraph{Measuring the reach of a stage.}
We restrict the attention of one stage at a time to the tokens within a radius of two patches on the original grid, and measure what share of its attention benefit, the loss gain of unrestricted over self-only attention, the stage keeps. This is a transformer analogue of the effective receptive field~\citep{luo2016erf,raghu2021vit}: a stage that keeps most of the benefit works locally, and a stage that loses it depends on distant tokens and works globally. We use the SiT-B/2 models of Section~\ref{sec:exp-b2} at 100K steps, with the training graph, noise level $t=0.5$ and three probe seeds (Appendix~\ref{app:probes}).

\begin{figure}[t]
  \centering
  \includegraphics[width=\linewidth]{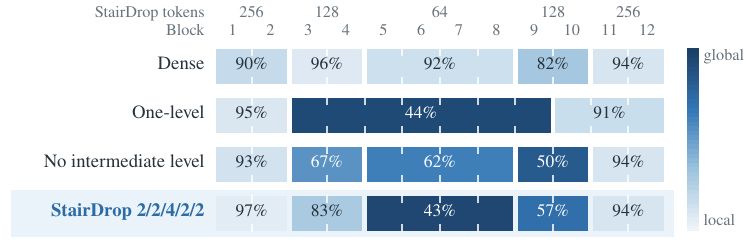}
  \caption{\textbf{How far each stage reaches.} SiT-B/2 at 100K steps, $t=0.5$. Each row is a model and each square a block. The color and the number give the share of a stage's attention benefit kept when its attention is restricted to a radius of two patches; darker stages depend more on distant tokens. Means over three probe seeds.}
  \label{fig:understanding}
\end{figure}

\subsection{One level: a bottleneck reached in a single step}
\label{sec:und-onelevel}

In the one-level model, the sparse span keeps 44\% of the benefit, while the same blocks of the dense model keep between 82\% and 96\% (Figure~\ref{fig:understanding}): the span works globally. The blocks after it read its output locally. When their direct attention to the kept outputs of the span is restricted to a radius of one patch, they still keep 96\% of the loss benefit of that output (Appendix~\ref{app:probes}). The sparse span therefore acts as the bottleneck of a U-Net, and one-level dropping reaches it in a single step: the kept share drops from 95\% to 44\% at the first sparse block and returns to 91\% after the span.

\subsection{The staircase: a bottleneck reached step by step}
\label{sec:und-depth}

With the intermediate level, the reach widens as participation falls (kept share from 97\% to 83\% and then 43\% in the core) and narrows again as positions are restored (from 57\% to 94\%), as along the contracting and expanding paths of a U-Net. The control with the same blocks and fusions but no intermediate level does not have this profile: its blocks 3--4 are already global (67\%), and the reach stays global through block 10.

\subsection{Why the intermediate level helps}
\label{sec:und-reading}

We read the reach as a balance between global structure and local detail. Widening it in a single step, as one-level dropping does, favors global structure at the expense of local detail. With the intermediate level, the blocks at 128 tokens still work locally, so the tokens that continue into the core first gather the detail of their neighbors, in two local steps rather than one, and the core then works globally on tokens that carry more local detail. The control removes this step-by-step profile, and with it the gain of Figure~\ref{fig:layouts}A, from 37.85 to 29.45.

\section{Conclusion}
\label{sec:conclusion}

\method{} schedules token participation along depth as a staircase: during training it lowers the number of participating tokens level by level into a sparse core and restores it level by level, and the network that generates images remains the dense backbone. The intermediate level improves on one-level dropping with the same blocks and parameters on SiT-B/2, and on one-level dropping on other backbones and in text-to-image generation, and our guidelines for dividing depth among the levels carry over from SiT-B/2 to SiT-XL/2. As a complete system, \method{} matches the FID that \sprint{} reports with classifier-free guidance using half of its training compute, and reaches FID 1.47 at the same compute. Inside the trained models, the staircase widens and narrows the reach of the network step by step, as a U-Net does with resolution.

\section{Limitations and Future Work}
\label{sec:limitations}

We study two levels with equal halving steps; on the twelve blocks of SiT-B/2, a third level inside the core does not improve on two. Our guidelines for dividing depth come from SiT-B/2 and are tested on SiT-XL/2. Which staircase is optimal for larger backbones, larger token grids and longer training, with how many levels, at which ratios and with how much depth each, remains open. Tokens are selected at random within each cell, and selection that depends on the image content is a natural extension. Large text-to-image systems such as Stable Diffusion~3~\citep{esser2024sd3} and FLUX~\citep{bfl2024flux} process image and text tokens jointly, and carrying the staircase to them, beyond our MS-COCO study at B/2 scale, is future work, as are higher resolutions and video, whose longer token sequences leave room for more levels.

\clearpage
\bibliography{references}
\bibliographystyle{iclr2027_conference}

\clearpage
\appendix
\section{Training and evaluation protocol}
\label{app:protocol}

\paragraph{Backbones.}
SiT-B/2 has 12 blocks of width 768 with 12 heads, and SiT-XL/2 has 28 blocks of width 1152 with 16 heads. Both use patch size 2 on the $32{\times}32{\times}4$ latents of \texttt{sd-vae-ft-ema}, which gives 256 tokens. The original backbone uses absolute sinusoidal position embeddings and uniform timestep sampling. The improved backbone used for Table~\ref{tab:system} is the configuration on which \sprint{} reports all of its numbers. It adds RMS normalization of the per-head queries and keys~\citep{dehghani2023vit22b,esser2024sd3}, two-dimensional rotary position embeddings~\citep{lu2024fit} that follow the kept token subset at every level, and logit-normal timestep sampling~\citep{esser2024sd3}.

\paragraph{Pre-training.}
All models use global batch 256, AdamW with learning rate $10^{-4}$ and no weight decay, bf16 autocast, EMA decay 0.9999, the linear interpolant with velocity target, and gradient clipping at norm 1.0. The fusions use Xavier initialization with zero bias, and the placeholders are fixed. Relative to one-level dropping with one fusion, \method{} adds $2D^2 + D$ learned parameters (2.7M on SiT-XL/2) and no Transformer block. The token masks are drawn per sample and per step from a dedicated generator that is checkpointed with the model, so a resumed run is bitwise identical to an uninterrupted one. Withheld positions carry the same loss weight as kept ones. Every SiT-B/2 arm is trained from scratch on the same latents with the same seed. The pipeline is deterministic, with two independent runs of one configuration agreeing to $10^{-7}$, so the arms differ only in their layout. Training compute follows \sprint{}'s convention, the forward FLOPs of the training graph at batch 256 times the number of iterations, measured with a FLOP counter (Appendix~\ref{app:compute}).

On SiT-B/2, \tread{} runs at its published setting for this depth, routing 50\% of the tokens through an identity path from block 3 to block 9. The one-level reference applies \sprint{}'s mechanism over the same blocks: blocks 3 to 9 run on one token per $2{\times}2$ cell, followed by placeholder restoration and one fusion. \sprint{}'s own SiT-B/2 setting spans blocks 3 to 10, which is the span of the one-level control with two fusions. On SiT-XL/2, \method{} $4/6/8/6/4$ runs blocks 5--10 and 19--24 on 128 tokens and blocks 11--18 on 64.

\paragraph{Dense stage.}
Before guided sampling, the sparse weights are trained for 20K steps with every position kept, at global batch 1024 with a 2.5K-step warm-up. The learning rate is $2 \times 10^{-4}$ for the 800K-step row, which is \sprint{}'s fine-tuning setting, and $10^{-4}$ for the 1.72M-step row. Everything else is unchanged. Unguided results use the sparse weights directly.

\paragraph{Sampling and scoring.}
SDE-250 is the Euler--Maruyama sampler of SiT with 250 steps, and ODE-50 is the Euler ODE sampler with 50 steps. Guidance follows Eq.~(\ref{eq:cfg}). It is applied along the whole trajectory for the 800K-step row. For the rows marked $\ddagger$, it is applied only for interpolation times $t \le 0.7$, the 70\% of the trajectory closest to the data. The guidance scale is chosen on 10K samples (Table~\ref{tab:cfg-sweep} for the two rows without REPA), and the reported 50K number is a single run at that scale: $w = 1.2$, $1.6$ and $1.6$ for the 800K-step, 1.72M-step and REPA rows. All samples are scored with the unmodified \texttt{evaluator.py} of guided-diffusion, the ADM evaluator, against the ImageNet $256{\times}256$ reference batch and its stored statistics, on 10K or 50K samples as stated. Every reported number uses EMA weights and sampling seed 0.

\section{Compute accounting}
\label{app:compute}

Training compute follows \sprint{}'s convention: the forward FLOPs of the training graph at batch 256, multiplied by the number of iterations. Published results use the configurations reported by their papers. Where available, their training compute is taken from \sprint{}'s measurements and scaled to those training lengths, and otherwise it is left unreported.

Epochs in Table~\ref{tab:system} follow the source papers or their tabulation in \sprint{}, with the following accounting of training stages. For our models and the \sprint{} rows, epochs are images seen divided by 1.28M, with every stage counted at its batch size. The \sprint{} rows use 1.9M sparse iterations at batch 256 plus 100K dense iterations at batch 1024, which gives 460 epochs, including the dense stage that \sprint{}'s own epoch count omits. MaskDiT's 1600 epochs cover masked training only, because the batch size of its 75K-step unmasking stage is not specified. For MDTv2~\citep{gao2024mdtv2}, the paper's own training length, 4.6M iterations at batch 256, replaces the 1080 epochs of \sprint{}'s table, and its compute is \sprint{}'s figure scaled to that length, $258.3 \times 4.6/5.4 = 220.0$.

For the additional backbone rows, ADM-G,U~\citep{dhariwal2021adm}, LDM-4~\citep{rombach2022ldm}, U-ViT-H/2~\citep{bao2023uvit} and FiTv2-XL/2~\citep{wang2024fitv2}, we report the metrics that the original papers give alongside their FID. ADM's 673M parameters include its base model, upsampler and guidance classifier, and we assign no single epoch count to this multistage system. U-ViT-H/2 trains for 500K steps at batch 1024, which corresponds to 400 epochs, and FiTv2 for 2M steps at batch 256, also 400 epochs. These rows keep their original samplers rather than our SDE-250 protocol.

A dense forward pass of our SiT-XL/2 model, including its two fusion layers, is 240 GFLOPs per sample (237.2 for the bare backbone), or 61.4 TFLOPs per batch of 256, measured with \texttt{torch.utils.flop\_counter}. This agrees with the 61 TFLOPs per iteration implied by \sprint{}'s Table~1. The $4/6/8/6/4$ training graph measures 136.5 GFLOPs per sample, or 34.9 TFLOPs per step, a ratio of 0.569 to dense, against 0.571 from the token-weighted block count of Eq.~(\ref{eq:cost}). The layouts $2/2/20/2/2$ and $3/3/16/3/3$ are counted by their token-weighted block counts, 0.393 and 0.464.

The dense stage costs $20\mathrm{K} \times 4 \times 61.4 = 4.9 \times 10^{6}$ TFLOPs. The \method{} rows of Table~\ref{tab:system} therefore cost $800\mathrm{K} \times 34.9 + 4.9 \times 10^{6} = 32.8 \times 10^{6}$ and $1.72\mathrm{M} \times 34.9 + 4.9 \times 10^{6} = 64.9 \times 10^{6}$ TFLOPs. The REPA row also counts its alignment head, 2.1 GFLOPs per sample on the 128 tokens of the aligned block during sparse training and 4.2 on 256 tokens in the dense stage, which gives $400\mathrm{K} \times 35.5 + 5.0 \times 10^{6} = 19.2 \times 10^{6}$. The DINOv2 features are precomputed and, as in REPA, not counted. \sprint{}'s quality metrics and training compute are quoted as published. From its Tables~1 and~9, its 400-epoch row consists of 1.9M sparse steps at 21.3 TFLOPs plus 100K dense steps at batch 1024, so 38\% of its $65.1\times10^6$ TFLOPs are spent in the dense stage.

\section{Comparison at equal training compute}
\label{app:equal-compute}

Sections~\ref{sec:exp-b2} and~\ref{sec:exp-xl} compare models at the same number of training steps. Here the same runs are compared by training compute, counted as in Appendix~\ref{app:compute}.

On SiT-B/2, a training step of \method{} $2/2/4/2/2$ costs 26.4 GFLOPs per sample, against 25.2 for one-level dropping, 22.6 for the one-level control with two fusions, and 43.6 for dense training. Table~\ref{tab:b2-equal} compares the models of Figure~\ref{fig:layouts}A when each has used at least the training compute of \method{} at 100K steps. Dense training and \tread{} cost more per step than \method{}, so their 100K-step results in Table~\ref{tab:b2long} already use more compute. On the settings of Table~\ref{tab:backbones}, one-level dropping trained until it has used at least the training compute of \method{} at 100K steps reaches FID-10K 23.91, 28.97 and 23.74, against 19.74, 21.64 and 19.12 for \method{}.

\begin{table}[h]
  \centering\small
  \caption{\textbf{SiT-B/2 at equal training compute.} FID-10K, ODE-50, no guidance. Compute is relative to \method{} at 100K steps.}
  \label{tab:b2-equal}
  \begin{tabular}{@{}lrrr@{}}
    \toprule
    Model & Steps & Training compute & FID-10K \\
    \midrule
    One-level control, two fusions & 116.9K & 1.00 & 33.32 \\
    One-level (\sprint{}) & 125K & 1.19 & 32.45 \\
    \method{} $2/2/4/2/2$ & 100K & 1.00 & \textbf{29.45} \\
    \bottomrule
  \end{tabular}
\end{table}

On SiT-XL/2, Table~\ref{tab:xl-equal} gives the runs of Table~\ref{tab:xl400} with the training compute of each evaluated step, relative to one-level dropping at 400K steps, from the step costs of Eq.~(\ref{eq:cost}). With no more than the compute of one-level dropping at 400K steps, $2/2/20/2/2$ reaches 8.47 at 0.83 of it and $3/3/16/3/3$ reaches 8.46 at 0.98, against 8.75. The layout used at system level, $4/6/8/6/4$, costs 0.571 of a dense step. Its comparison at equal compute is the system-level one of Section~\ref{sec:exp-system}, with guidance and the dense stage.

\begin{table}[h]
  \centering\small
  \caption{\textbf{SiT-XL/2 runs by training compute.} FID-10K, ODE-50, no guidance, sparse weights; training compute relative to one-level dropping at 400K steps in parentheses.}
  \label{tab:xl-equal}
  \begin{tabular}{@{}lcccc@{}}
    \toprule
    Layout & Step cost & 200K & 300K & 400K \\
    \midrule
    One-level $2/{-}/24/{-}/2$ & 0.357 & 11.52 (0.50) & 9.65 (0.75) & 8.75 (1.00) \\
    \method{} $2/2/20/2/2$ & 0.393 & 9.74 (0.55) & 8.47 (0.83) & 8.08 (1.10) \\
    \method{} $3/3/16/3/3$ & 0.464 & 11.70 (0.65) & 8.46 (0.98) & 7.52 (1.30) \\
    \method{} $4/6/8/6/4$ & 0.571 & 11.20 (0.80) & 8.76 (1.20) & 7.68 (1.60) \\
    \bottomrule
  \end{tabular}
\end{table}

\section{Restoration variants}
\label{app:restore}

\begin{table}[ht]
  \centering\small
  \caption{\textbf{Restoration variants.} SiT-B/2 \method{} $2/2/3/2/3$ at 100K steps, FID-10K with SDE-250.}
  \label{tab:restore}
  \begin{tabular}{@{}lr@{}}
    \toprule
    Restore & FID-10K \\
    \midrule
    \method{}: placeholder at each withheld position, then fusion with the skip & \textbf{31.95} \\
    Span output appended as context tokens for the decoder; no restore, no fusion & 55.57 \\
    Placeholder replaced by a learned upsampler of the span's output & 49.15 \\
    Skip half of the fusion dropped with probability 0.1 during training & 47.50 \\
    \bottomrule
  \end{tabular}
\end{table}

Table~\ref{tab:restore} compares alternatives to the placeholder-and-skip restoration that \method{} takes from \sprint{}. Appending the output of the span as context tokens changes both the restoration and the fusion. The other two variants change the placeholder or regularize the skip. All three are worse in this setting, and \method{} keeps the original restoration.

\section{Probe protocol}
\label{app:probes}

The curves of Table~\ref{tab:reach} use the dense and one-level (\sprint{}) SiT-B/2 EMA checkpoints at 100K steps from Table~\ref{tab:b2full}. We evaluate the training graph on 2048 ImageNet training latents per probe seed, with class labels, no guidance and the fixed noise level $t=0.5$. The three probe sampling seeds vary the images, the latent and noise draws and the token masks, not the training run. Each draw is replayed across all restrictions. Radii are Chebyshev distances on the original $16{\times}16$ token grid, so $d=15$ is unrestricted.

Figure~\ref{fig:understanding} applies the within-span probe below to each stage of the dense model, the one-level model, the control without the intermediate level and \method{} $2/2/4/2/2$ at 100K steps, one stage at a time, at $d=2$ and $t=0.5$ over the same three probe seeds. Section~\ref{sec:und-onelevel} uses the after-restoration probe on the one-level model, at $d=1$ and $t=0.5$ (Table~\ref{tab:reach}).

\paragraph{Within the span.}
In blocks 3--9, each query attends only to the active positions within radius $d$, itself included. With $\ell_d$ the resulting velocity loss, we report the normalized attention benefit
\begin{equation}
  C(d)=\frac{\ell_0-\ell_d}{\ell_0-\ell_{15}} .
  \label{eq:probe-computed-share}
\end{equation}
The reference $\ell_0$ uses self-only attention, with the residual paths and MLPs still active. $C(d)$ therefore measures the benefit of attention across positions relative to this reference, which is a narrower quantity than the contribution of the whole span. The dense reference applies the same restriction to the same block indices, with all positions active.

\paragraph{After restoration.}
In blocks 10--12 of the one-level model, queries may attend directly to the positions kept through the span only within radius $d$, while attention to the restored positions stays unrestricted. Let $\ell_d^{\mathrm{out}}$ be the loss with the span output present and $\ell_d^{\mathrm{ref}}$ the loss when that output is replaced by the fixed placeholder before fusion, with the skip and the same attention restriction kept. The normalized output benefit is
\begin{equation}
  R(d)=\frac{\ell_d^{\mathrm{ref}}-\ell_d^{\mathrm{out}}}
                 {\ell_{15}^{\mathrm{ref}}-\ell_{15}^{\mathrm{out}}} .
  \label{eq:probe-read-share}
\end{equation}
The restriction applies to direct attention only, and restored positions can still relay information from the kept outputs across the later blocks. $C(d)$ and $R(d)$ have different reference losses. Both are normalized contrasts, both can exceed one, and both are computed within each seed before averaging. Table~\ref{tab:reach} gives the full radius sweep of both probes at all three measured noise levels.

\begin{table}[b]
  \centering\small
  \caption{\textbf{Motivation probes across noise levels.} Mean normalized loss benefits from Eqs.~(\ref{eq:probe-computed-share}) and~(\ref{eq:probe-read-share}) over three probe sampling seeds. Radii are in patch units of the original grid.}
  \label{tab:reach}
  \setlength{\tabcolsep}{4pt}
  \begin{tabular}{@{}llrrrrrrrr@{}}
    \toprule
    Probe / model & $t$ & $d=0$ & $1$ & $2$ & $3$ & $4$ & $6$ & $8$ & $15$ \\
    \midrule
    \multirow{3}{*}{$C(d)$: Dense}
      & 0.1 & 0.00 & 0.80 & 0.91 & 0.95 & 0.97 & 0.98 & 0.99 & 1.00 \\
      & 0.5 & 0.00 & 0.71 & 0.87 & 0.94 & 0.97 & 0.99 & 0.99 & 1.00 \\
      & 0.9 & 0.00 & 0.84 & 0.94 & 0.98 & 0.99 & 1.00 & 1.00 & 1.00 \\
    \addlinespace[3pt]
    \multirow{3}{*}{$C(d)$: One-level}
      & 0.1 & 0.00 & 0.25 & 0.56 & 0.71 & 0.80 & 0.88 & 0.93 & 1.00 \\
      & 0.5 & 0.00 & 0.17 & 0.44 & 0.60 & 0.71 & 0.84 & 0.91 & 1.00 \\
      & 0.9 & 0.00 & 0.06 & 0.21 & 0.37 & 0.51 & 0.71 & 0.83 & 1.00 \\
    \addlinespace[3pt]
    \multirow{3}{*}{$R(d)$: One-level}
      & 0.1 & 0.14 & 0.83 & 0.93 & 0.95 & 0.97 & 0.98 & 0.99 & 1.00 \\
      & 0.5 & 0.09 & 0.96 & 1.02 & 1.03 & 1.02 & 1.02 & 1.01 & 1.00 \\
      & 0.9 & 0.09 & 0.81 & 0.90 & 0.94 & 0.97 & 0.99 & 1.00 & 1.00 \\
    \bottomrule
  \end{tabular}
\end{table}

\clearpage
\section{Full metrics}
\label{app:full}

This appendix collects the complete SiT-B/2 design study, the training trajectories of the default layouts and of the three SiT-XL/2 layouts, and the guidance-scale sweep behind the system-level rows.

\begin{table}[ht]
  \centering\small
  \caption{\textbf{Complete SiT-B/2 design study.} ODE-50, no guidance; 10K samples at 100K steps and 50K samples at 400K steps. The one-level rows keep 64 tokens through blocks 3--9 (one fusion) or 3--10 (two fusions).}
  \label{tab:b2full}
  \begin{tabular}{@{}lrrr@{}}
    \toprule
    & \multicolumn{2}{c}{100K steps} & 400K steps \\
    \cmidrule(lr){2-3}\cmidrule(lr){4-4}
    Model & FID-10K & IS & FID-50K \\
    \midrule
    Dense & 70.72 & 20.0 & 38.58 \\
    \tread{} (route 50\%, blocks 3--9) & 54.71 & 28.1 & 20.58 \\
    One-level, one fusion (blocks 3--9) & 39.25 & 41.7 & 11.48 \\
    One-level, two fusions & 37.85 & 42.4 & \\
    \method{} $2/2/4/2/2$ & \textbf{29.45} & \textbf{53.0} & \textbf{7.66} \\
    \method{} $2/1/6/1/2$ & 30.91 & 50.4 & \\
    \method{} $1/2/6/2/1$ & 33.74 & 44.5 & \\
    \method{} $3/2/4/2/1$ & 35.11 & 43.0 & \\
    \method{} $2/3/2/3/2$ & 36.41 & 40.4 & \\
    \method{} $1/2/4/2/3$ & 38.48 & 40.6 & \\
    \midrule
    \method{} $2/2/3/2/3$, first level at 128 tokens & 33.52 & 45.4 & \\
    \method{} $2/2/3/2/3$, first level at 192 tokens & 36.93 & 41.4 & \\
    \bottomrule
  \end{tabular}
\end{table}

\begin{table}[ht]
  \centering\small
  \caption{\textbf{SiT-B/2 \method{} $2/2/4/2/2$ along training.} SDE-250, 50K samples, no guidance.}
  \label{tab:b2-sde}
  \begin{tabular}{@{}lrrrrr@{}}
    \toprule
    Steps & FID & sFID & IS & Prec. & Rec. \\
    \midrule
    100K & 24.89 & 9.08 & 54.4 & 0.653 & 0.459 \\
    200K & 12.25 & 7.44 & 107.0 & 0.761 & 0.452 \\
    300K & 8.72 & 7.02 & 137.8 & 0.791 & 0.473 \\
    400K & 7.18 & 6.91 & 160.5 & 0.807 & 0.463 \\
    \bottomrule
  \end{tabular}
\end{table}

\begin{table}[ht]
  \centering\small
  \caption{\textbf{SiT-XL/2 \method{} $4/6/8/6/4$ along training.} Original backbone, sparse weights; ODE-50, 50K samples, no guidance.}
  \label{tab:xl-long}
  \begin{tabular}{@{}lrrrrr@{}}
    \toprule
    Steps & FID & sFID & IS & Prec. & Rec. \\
    \midrule
    400K & 4.89 & 4.77 & 173.8 & 0.816 & 0.520 \\
    500K & 4.40 & 4.68 & 185.3 & 0.820 & 0.515 \\
    600K & 4.02 & 4.66 & 193.0 & 0.823 & 0.530 \\
    700K & 3.78 & 4.64 & 200.1 & 0.823 & 0.528 \\
    800K & 3.61 & 4.57 & 205.5 & 0.827 & 0.543 \\
    \bottomrule
  \end{tabular}
\end{table}

\begin{table}[ht]
  \centering\small
  \caption{\textbf{The three SiT-XL/2 layouts of Table~\ref{tab:xl400} along training.} ODE-50, FID-10K, no guidance.}
  \label{tab:xl-layouts}
  \begin{tabular}{@{}lrrrr@{}}
    \toprule
    Layout & 100K & 200K & 300K & 400K \\
    \midrule
    $2/2/20/2/2$ & 17.03 & 9.74 & 8.47 & 8.08 \\
    $3/3/16/3/3$ & 25.07 & 11.70 & 8.46 & 7.52 \\
    $4/6/8/6/4$ & 22.41 & 11.20 & 8.76 & 7.68 \\
    \bottomrule
  \end{tabular}
\end{table}

\begin{table}[ht]
  \centering\small
  \caption{\textbf{Guidance-scale sweep for the two \method{} rows of Table~\ref{tab:system} without REPA.} SDE-250, FID-10K. Interval guidance is applied only for $t \le 0.7$.}
  \label{tab:cfg-sweep}
  \begin{tabular}{@{}lrrrrrrr@{}}
    \toprule
    Model & $w{=}1.2$ & $1.3$ & $1.4$ & $1.5$ & $1.6$ & $1.7$ & $1.8$ \\
    \midrule
    800K + 20K, whole trajectory & \textbf{4.98} & 5.21 & -- & -- & -- & -- & -- \\
    800K + 20K, interval & -- & -- & 4.74 & \textbf{4.73} & -- & -- & -- \\
    1.72M + 20K, interval & -- & -- & 4.77 & 4.58 & \textbf{4.53} & 4.61 & 4.74 \\
    \bottomrule
  \end{tabular}
\end{table}

\FloatBarrier
\section{Generated samples}
\label{app:qualitative}

The figures below show further selected samples from the same evaluation set as Figure~\ref{fig:qualitative-main}, generated by the SiT-XL/2 model with FID-50K 1.47. Each category has 27 distinct $256{\times}256$ images, arranged in rows of 3, 6, 9 and 9 at different display sizes.

\begin{figure}[!ht]
  \centering
  \includegraphics[width=0.92\linewidth]{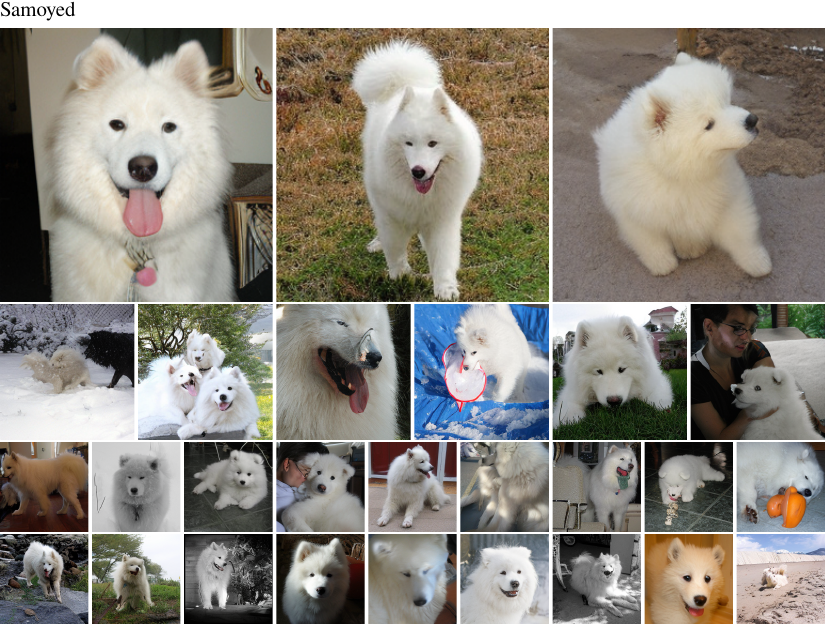}\\[0.6em]
  \includegraphics[width=0.92\linewidth]{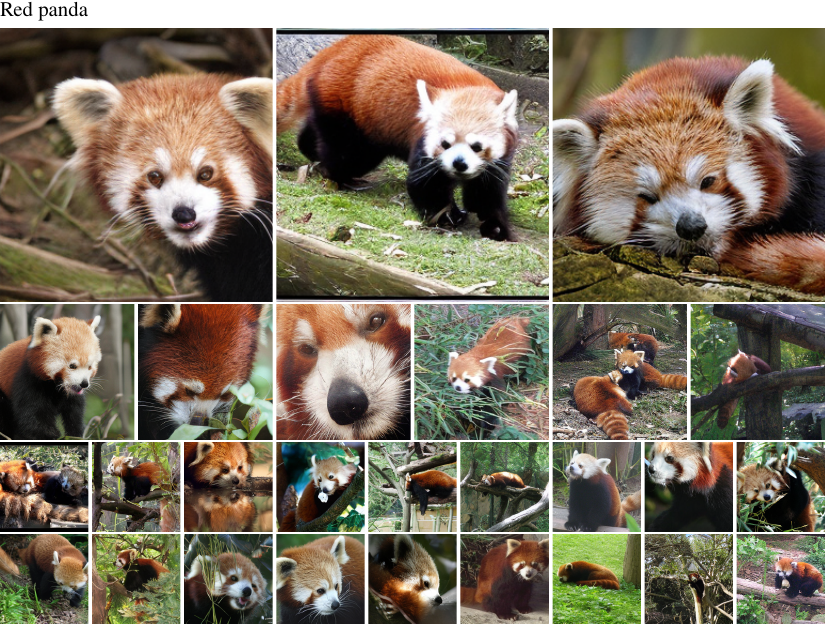}
  \caption{\textbf{Selected \method{} samples.} Samoyed and red panda.}
  \label{fig:qualitative-animals}
\end{figure}

\clearpage
\begin{figure}[!ht]
  \centering
  \includegraphics[width=0.92\linewidth]{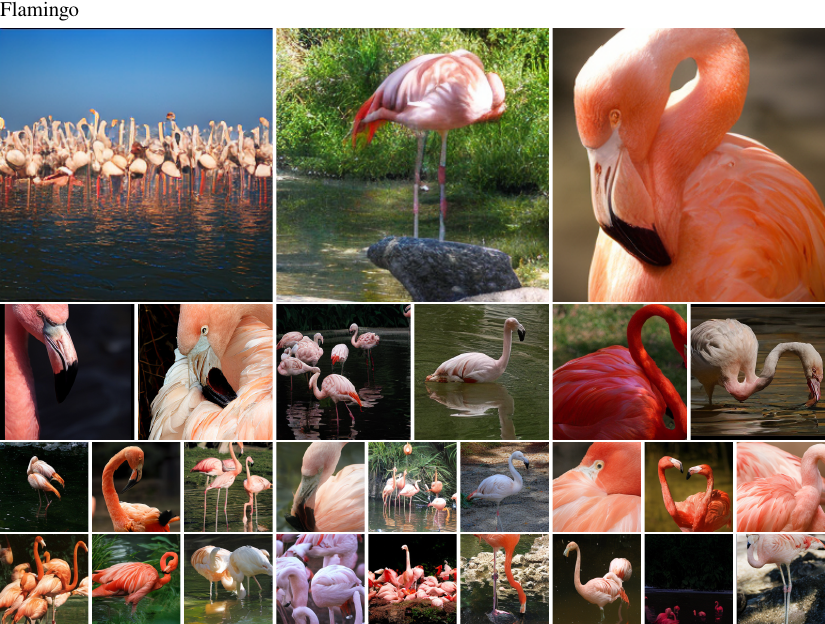}\\[0.6em]
  \includegraphics[width=0.92\linewidth]{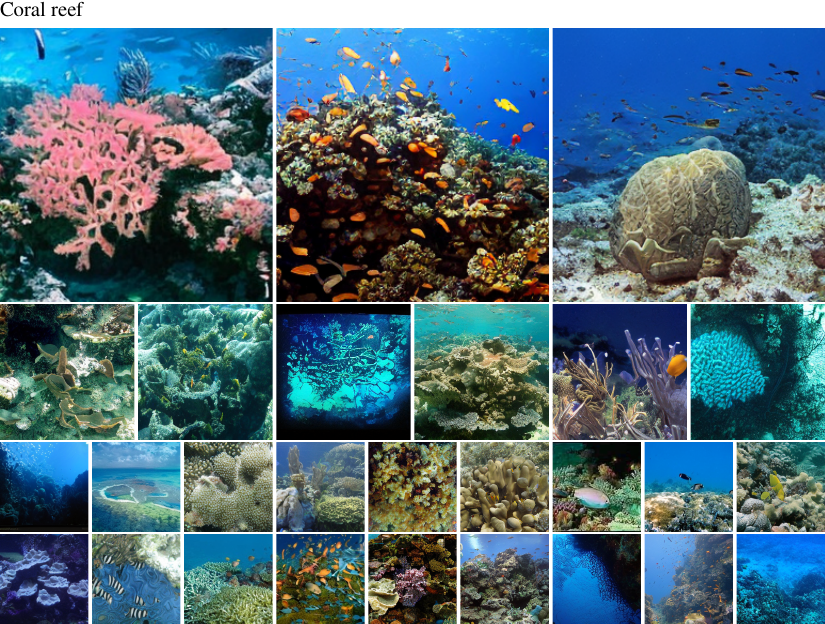}
  \caption{\textbf{Selected \method{} samples.} Flamingo and coral reef.}
  \label{fig:qualitative-nature}
\end{figure}

\clearpage
\begin{figure}[!ht]
  \centering
  \includegraphics[width=0.92\linewidth]{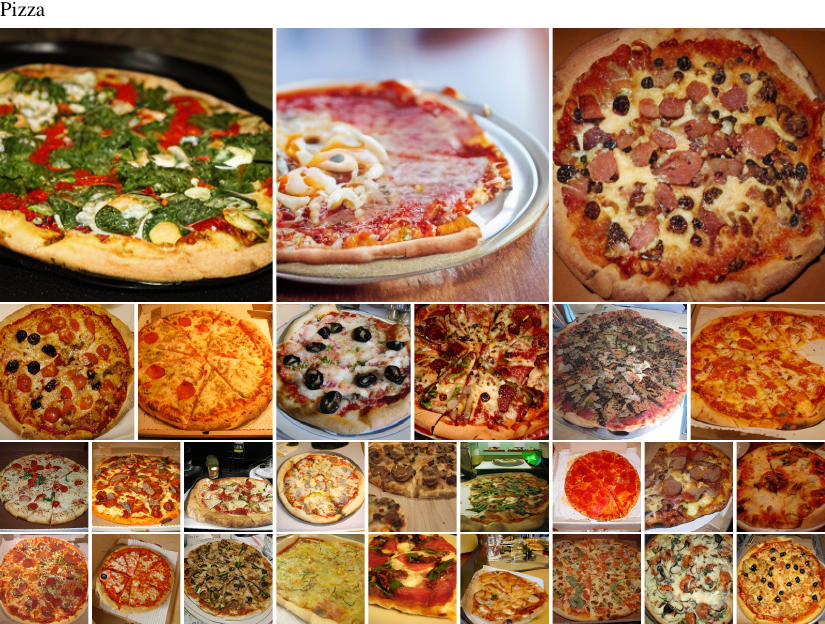}\\[0.6em]
  \includegraphics[width=0.92\linewidth]{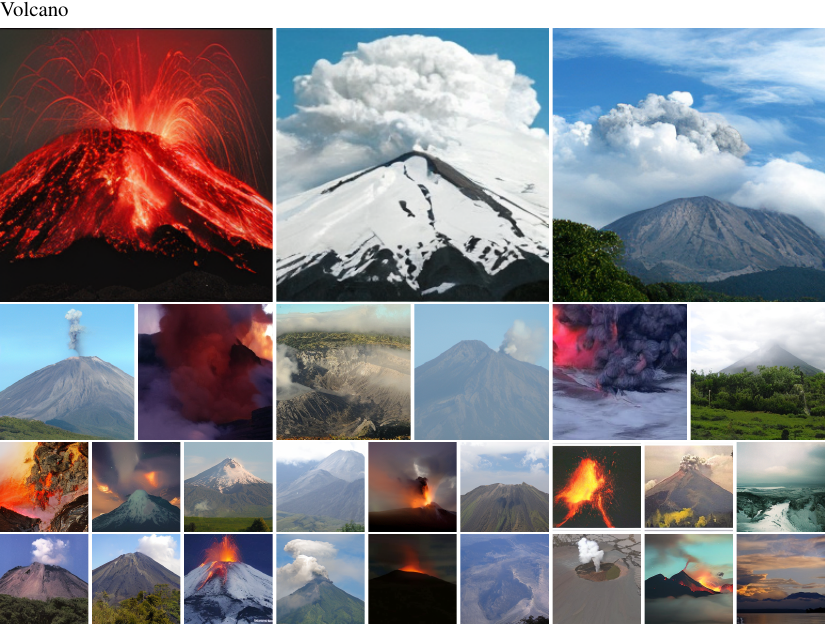}
  \caption{\textbf{Selected \method{} samples.} Pizza and volcano.}
  \label{fig:qualitative-food-landscape}
\end{figure}

\end{document}